\documentclass[twocolumn,switch,times]{article}
\usepackage{preprint}
\usepackage{hyperref}
\usepackage[authoryear,round]{natbib}

\hypersetup{
    colorlinks=true,
    linkcolor=blue,
    urlcolor=blue,
    citecolor=blue,
    anchorcolor=black
}

\usepackage[utf8]{inputenc}	
\usepackage[T1]{fontenc}
\usepackage{xspace}
\usepackage{lineno}					
\usepackage{tikz} 					
\usepackage{fancyhdr}
\usepackage{geometry}
\usepackage{lipsum}
\usepackage{scalerel}
\usepackage{sidecap}
\usepackage{times}
\usepackage{doi}
\usepackage{amssymb}
\usepackage{amsmath}
\usepackage{bbm}
\usepackage{bm}
\usepackage{array,booktabs}
\usepackage{upgreek}
\usepackage{algorithm}
\usepackage{algpseudocodex}
\usepackage{appendix}
\usepackage{authblk}
\usepackage{placeins}

\usepackage{caption}
\usepackage{subcaption}
\usepackage{multirow}
\usepackage{pifont}
\usepackage{longtable}
\usepackage{array}
\usepackage{booktabs}
\usepackage{multirow}
\usepackage{float}
\usepackage{siunitx}
\usepackage{caption}
\usepackage{makecell}

\usepackage[utf8]{inputenc}
\usepackage{listings}
\lstdefinestyle{promptstyle}{
    backgroundcolor=\color{black!5},   
    frame=single,                       
    framerule=0.5pt,                    
    rulecolor=\color{black!60},         
    breaklines=true,                    
    basicstyle=\ttfamily\small,         
    keywordstyle=\color{blue},          
    commentstyle=\color{green!60!black},
    stringstyle=\color{purple},         
    captionpos=b,                       
    abovecaptionskip=5pt,               
    belowcaptionskip=5pt,               
}
\definecolor{codegreen}{rgb}{0,0.6,0}
\definecolor{codegray}{rgb}{0.5,0.5,0.5}
\definecolor{codepurple}{rgb}{0.58,0,0.82}
\definecolor{backcolour}{rgb}{0.95,0.95,0.92}

\lstdefinestyle{codestyle}{
    language=Python, 
    backgroundcolor=\color{backcolour},   
    commentstyle=\color{codegreen},
    keywordstyle=\color{magenta},
    numberstyle=\tiny\color{codegray},
    stringstyle=\color{codepurple},
    basicstyle=\ttfamily\footnotesize,
    breakatwhitespace=false,         
    breaklines=true,                 
    captionpos=b,                    
    keepspaces=true,                 
    numbers=left,                    
    numbersep=5pt,                  
    showspaces=false,                
    showstringspaces=false,
    showtabs=false,                  
    tabsize=2,
    frame=single, 
    rulecolor=\color{black!30}, 
    framerule=0.5pt, 
    title=\lstname 
}
\usepackage{longtable}
\usepackage{array}
\usepackage{booktabs}
\usepackage{graphicx}
\usepackage{multirow}
\usepackage{float}
\usepackage{siunitx}
\usepackage{caption}

\usepackage{newfloat}
\DeclareFloatingEnvironment[name={Supplementary Figure}]{suppfigure}
\usepackage{sidecap}
\sidecaptionvpos{figure}{c}
\usepackage{etoolbox}

\usepackage{titlesec}
\titlespacing\section{0pt}{12pt plus 3pt minus 3pt}{1pt plus 1pt minus 1pt}
\titlespacing\subsection{0pt}{10pt plus 3pt minus 3pt}{1pt plus 1pt minus 1pt}
\titlespacing\subsubsection{0pt}{8pt plus 3pt minus 3pt}{1pt plus 1pt minus 1pt}

\definecolor{lime}{HTML}{A6CE39}

\usepackage[table]{xcolor}
\usepackage[dvipsnames, table]{xcolor}
\usepackage[export]{adjustbox}

\definecolor{pltblue}{RGB}{174, 199, 232}
\definecolor{pltorange}{RGB}{255, 229, 204}
\definecolor{pltgreen}{RGB}{204, 229, 204}
\definecolor{pltred}{RGB}{229, 204, 204}
\definecolor{pltpurple}{RGB}{239, 218, 230}

\definecolor{tabblue}{HTML}{1f77b4}
\definecolor{taborange}{HTML}{ff7f0e}
\definecolor{tabgreen}{HTML}{2ca02c}
\definecolor{tabred}{HTML}{d62728}
\definecolor{tabpurple}{HTML}{9467bd}
\definecolor{tabpink}{HTML}{ff0080}

\definecolor{cblue}{RGB}{173, 201, 233}
\definecolor{clblue}{RGB}{222, 234, 246}
\definecolor{corange}{RGB}{255, 152, 67}
\definecolor{lorgange}{RGB}{255, 221, 149}
\definecolor{tablegray}{RGB}{180, 180, 180}

\definecolor{cvprpink}{rgb}{0.858, 0.188, 0.478}
\newcommand{\samov}{SegEarth-OV\textcolor{cvprpink}{3}\xspace}

\newcolumntype{P}[1]{>{\centering\arraybackslash}p{#1}}

\newcommand{\appendixref}[1]{%
  \hyperref[#1]{Appendix~\ref*{#1}}\unskip
}

\title{\samov: Exploring SAM 3 for Open-Vocabulary Semantic Segmentation in Remote Sensing Images}

\usepackage{authblk}

\author[1,\small\ensuremath{\ast}]{Kaiyu Li}
\author[2,\small\ensuremath{\ast}]{Shengqi Zhang}
\author[2,\small\ensuremath{\ast}]{Yujie Wang}
\author[3]{Yupeng Deng}
\author[1]{Zhi Wang}
\author[4,5,6]{Deyu Meng}
\author[2,5,$\dagger$]{Xiangyong Cao}

\affil[1]{School of Software Engineering, Xi’an Jiaotong University, Xi’an, China}
\affil[2]{School of Computer Science and Technology, Xi’an Jiaotong University, Xi’an, China}
\affil[3]{School of Information and Communications Engineering, Xi’an Jiaotong University, Xi’an, China}
\affil[4]{School of Mathematics and Statistics, Xi’an Jiaotong University, Xi’an, China}
\affil[5]{Ministry of Education Key Lab of Intelligent Network Security, Xi’an, China}
\affil[6]{Pazhou Laboratory (Huangpu), Radarweg 29, Guangzhou, China}

\begin{document}

\twocolumn[\begin{@twocolumnfalse}

\maketitle

\begin{abstract}
Most existing methods for training-free open-vocabulary semantic segmentation are based on CLIP. While these approaches have made progress, they often face challenges in precise localization or require complex pipelines to combine separate modules, especially in remote sensing scenarios where numerous dense and small targets are present. Recently, Segment Anything Model 3 (SAM 3) was proposed, unifying segmentation and recognition in a promptable framework. In this paper, we present a comprehensive exploration of applying SAM 3 to the remote sensing open-vocabulary tasks (\textit{i.e.}, 2D semantic segmentation, change detection, and 3D semantic segmentation) without any training. First, we implement a mask fusion strategy that combines the outputs from SAM 3's semantic segmentation head and the Transformer decoder (instance head). This allows us to leverage the strengths of both heads for better land coverage. Second, we utilize the presence score from the presence head to filter out categories that do not exist in the scene, reducing false positives caused by the vast vocabulary sizes and patch-level processing in geospatial scenes. Furthermore, we extend our method to open-vocabulary change detection by a joint instance- and pixel-level verification strategy built directly upon our fused logits. We evaluate our method on extensive remote sensing datasets and tasks, including 20 segmentation datasets, 3 change detection datasets, and a 3D segmentation dataset. Experiments show that our method achieves promising performance, demonstrating the potential of SAM 3 for remote sensing open-vocabulary tasks. Our code is released at \url{https://github.com/earth-insights/SegEarth-OV-3}.\\
\end{abstract}


\vspace{0.5cm}

\end{@twocolumnfalse}]

\renewcommand{\thefootnote}{*}
\footnotetext[1]{Equal contribution}
\renewcommand{\thefootnote}{\dag}
\footnotetext[2]{Corresponding author (caoxiangyong@mail.xjtu.edu.cn)}
\renewcommand{\thefootnote}{\arabic{footnote}}

\setcounter{footnote}{0}

\section{Introduction}
\label{sec:intro}

Semantic segmentation is a fundamental task in remote sensing analysis, enabling dense, pixel-level understanding of Earth observation scenes. Traditionally, segmentation models were restricted to a closed set of predefined categories, limiting their applicability in dynamic, open-world scenarios where visual concepts are virtually infinite. To overcome this limitation, Open-Vocabulary Semantic Segmentation (OVSS) has emerged as a critical research direction~\citep{li2025segearth}. By leveraging the rich semantic knowledge embedded in pre-trained Vision-Language Models (VLMs), remote sensing OVSS aims to segment and recognize image regions based on arbitrary text descriptions, effectively generalizing to categories unseen during training. This capability is essential for diverse applications, \textit{e.g.}, urban planning~\citep{li2024show} and disaster monitoring~\citep{wang2025disasterm3}, where the model must handle a vast vocabulary~\citep{li2025segearth, li2025annotation, li2026dynamicearth}.


Currently, the dominant paradigm for training-free OVSS relies heavily on VLMs, particularly CLIP~\citep{radford2021learning}. Early works, such as MaskCLIP~\citep{zhou2022extract} and SCLIP~\citep{wang2023sclip}, attempt to extract dense features directly from the CLIP image encoder. However, CLIP is pre-trained for image-level recognition, and adapting its patch-level features for pixel-level localization often results in coarse boundaries. To address this, subsequent research has focused on integrating auxiliary Visual Foundation Models (VFMs). For instance, ProxyCLIP~\citep{lan2025proxyclip} and CorrCLIP~\citep{zhang2025corrclip} utilize structural guidance from DINO~\citep{oquab2023dinov2} and SAM~\citep{kirillov2023segment} to refine CLIP's attention maps, while SegEarth-OV~\citep{li2025segearth, li2025annotation} builds an upsampler to reconstruct high-resolution features. 
Although these methods improve boundary quality, they rely on complex pipeline and feature alignment to bridge the gap between different representations. Moreover, they lack simultaneous semantic and instance segmentation capabilities, limiting their utility in complex geospatial analysis.

Recently, the Segment Anything Model 3 (SAM 3)~\citep{carion2025sam} was introduced. Unlike CLIP-based paradigm, SAM 3 is a unified model that supports promptable concept segmentation. It builds upon the DETR~\citep{carion2020end} and MaskFormer~\citep{cheng2021per, cheng2022masked} architectures, employing a query-based Transformer design. Crucially, SAM 3 utilizes a decoupled architecture in which a presence head is specifically designed to predict the probability that the prompted concept exists in the image. Meanwhile, a Transformer decoder and a semantic segmentation head generate precise masks for discrete instances and continuous semantic regions, respectively. Although SAM 3 demonstrates impressive zero-shot capabilities on many natural image semantic segmentation benchmarks, remote sensing images present distinct challenges, \textit{e.g.}, the intricate coexistence of dense small objects and vast amorphous backgrounds. Therefore, a tailored exploration to adapt SAM 3 for geospatial scenarios remains valuable.

In this paper, we present a preliminary exploration of adapting SAM 3 for the remote sensing OVSS and some derived tasks without additional training. We investigate whether SAM 3's unified architecture can offer a stronger, yet simpler baseline than complex CLIP-ensemble methods for Earth observation. Our proposed method, namely \samov, consists of two simple strategies tailored to the design of SAM 3 and the characteristics of remote sensing scenarios:

\begin{itemize}
    \item \textbf{(1) Dual-Head Mask Fusion:} Remote sensing images exhibit a distinct duality: amorphous ``stuff'' (\textit{e.g.}, road, bareland) requiring pixel-wise semantic continuity, and countable ``things'' (\textit{e.g.}, buildings, vehicles) demanding instance-level boundary precision. We identify that SAM 3's decoupled architecture naturally aligns with this duality. We propose to assign the semantic head to maintain land-cover completeness and the Transformer decoder to capture fine-grained instance details, thereby unifying these complementary strengths to ensure robust segmentation across diverse geospatial targets.
    \item \textbf{(2) Presence-Guided Filtering:} In remote sensing OVSS, a complete vocabulary list might include global land cover types, but a single image patch covers a minute geographical area (\textit{e.g.}, hundreds of meters). This results in a high category sparsity, where the vast majority of queried concepts are physically absent in the local view. We leverage the presence score to address this global-local gap, explicitly suppressing irrelevant categories to eliminate false positives caused by the vast vocabulary against limited visual content~\citep{sun2024clip, chen2025training}.
\end{itemize}

Furthermore, we extend our method to Open-Vocabulary Change Detection (OVCD). In the open-vocabulary setting, VLMs often struggle with appearance variations across bi-temporal images (\textit{e.g.}, seasonal shifts or illumination differences), leading to inconsistent recognition results. Moreover, directly comparing uncalibrated logits is highly sensitive to bi-temporal registration errors. To address this, we introduce a joint instance- and pixel-level verification strategy built on \samov. By combining instance-level mask overlap and pixel-level feature similarity, this strategy effectively filters out registration-induced noise and VLM recognition inconsistencies to identify true temporal changes.

We evaluate our method on an extensive suite of remote sensing benchmarks, including 20 semantic segmentation datasets, 3 change detection datasets, and a 3D point cloud segmentation dataset. Our results demonstrate the strong capability of SAM~3 for remote sensing OVSS, which is further enhanced by our proposed improvements.

\section{Related Work}
\label{sec:related_work}

\begin{figure}[t]
    \centering
    \begin{subfigure}[b]{0.45\textwidth}
        \centering
        \includegraphics[width=\linewidth]{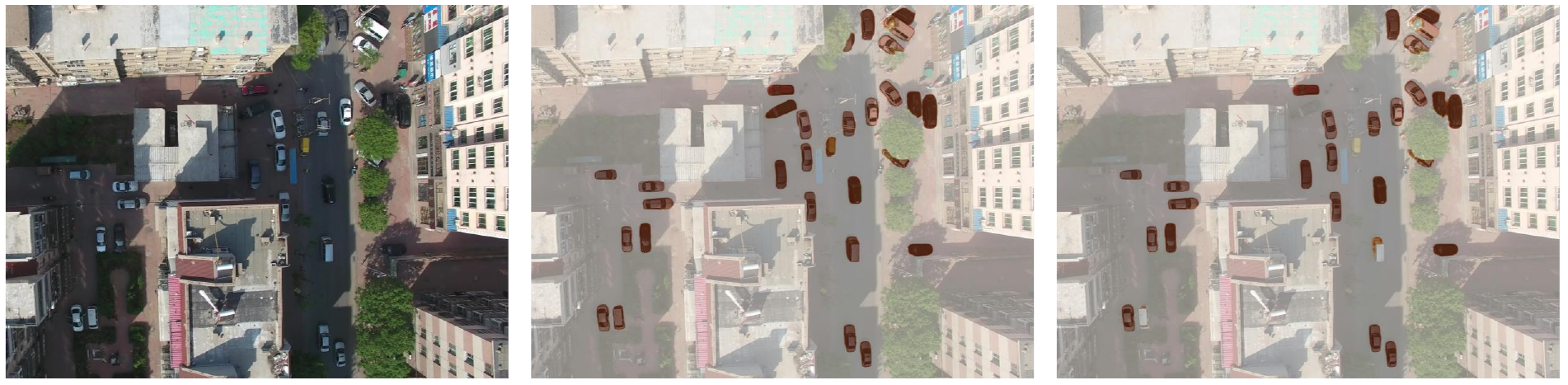} 
        \caption{Countable objects (\textit{e.g.}, cars).}
        \label{fig:mo1}
    \end{subfigure}
    \par 
    \begin{subfigure}[b]{0.45\textwidth}
        \centering
        \includegraphics[width=\linewidth]{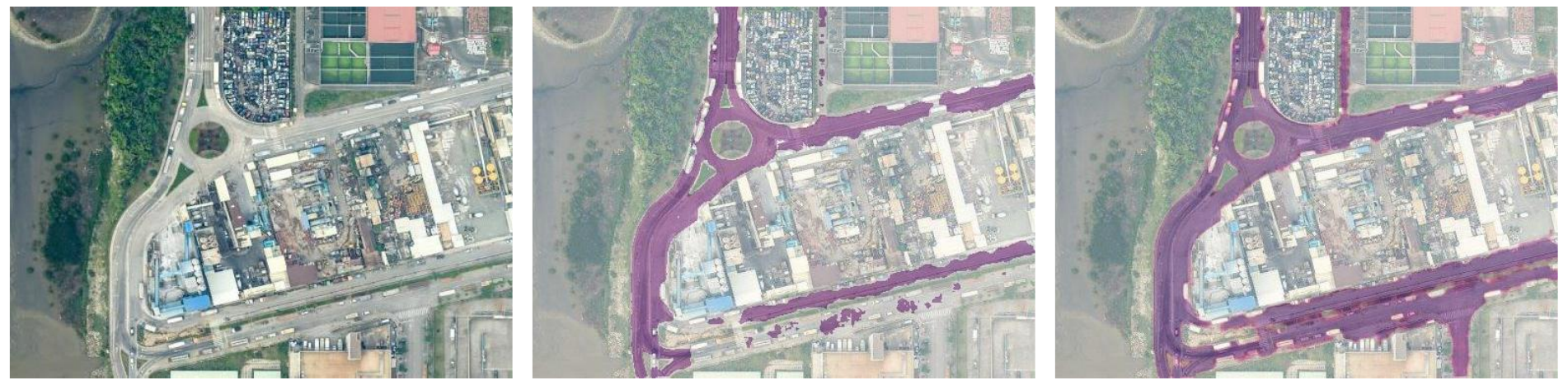}
        \caption{Amorphous regions (\textit{e.g.}, road).}
        \label{fig:mo2}
    \end{subfigure}
    \caption{The Transformer decoder of SAM 3 excels at delineating countable objects but produces fragmented masks for amorphous regions, while the semantic head preserves continuity for amorphous regions but lacks boundary precision for small targets. (Left: Remote sensing image. Middle: Prediction of Transformer Decoder. Right: Prediction of semantic segmentation head.)}
    \label{fig:motivation}
    \vspace{-1em}
\end{figure}

\begin{figure*}[t]
  \centering
   \includegraphics[width=1.0\linewidth]{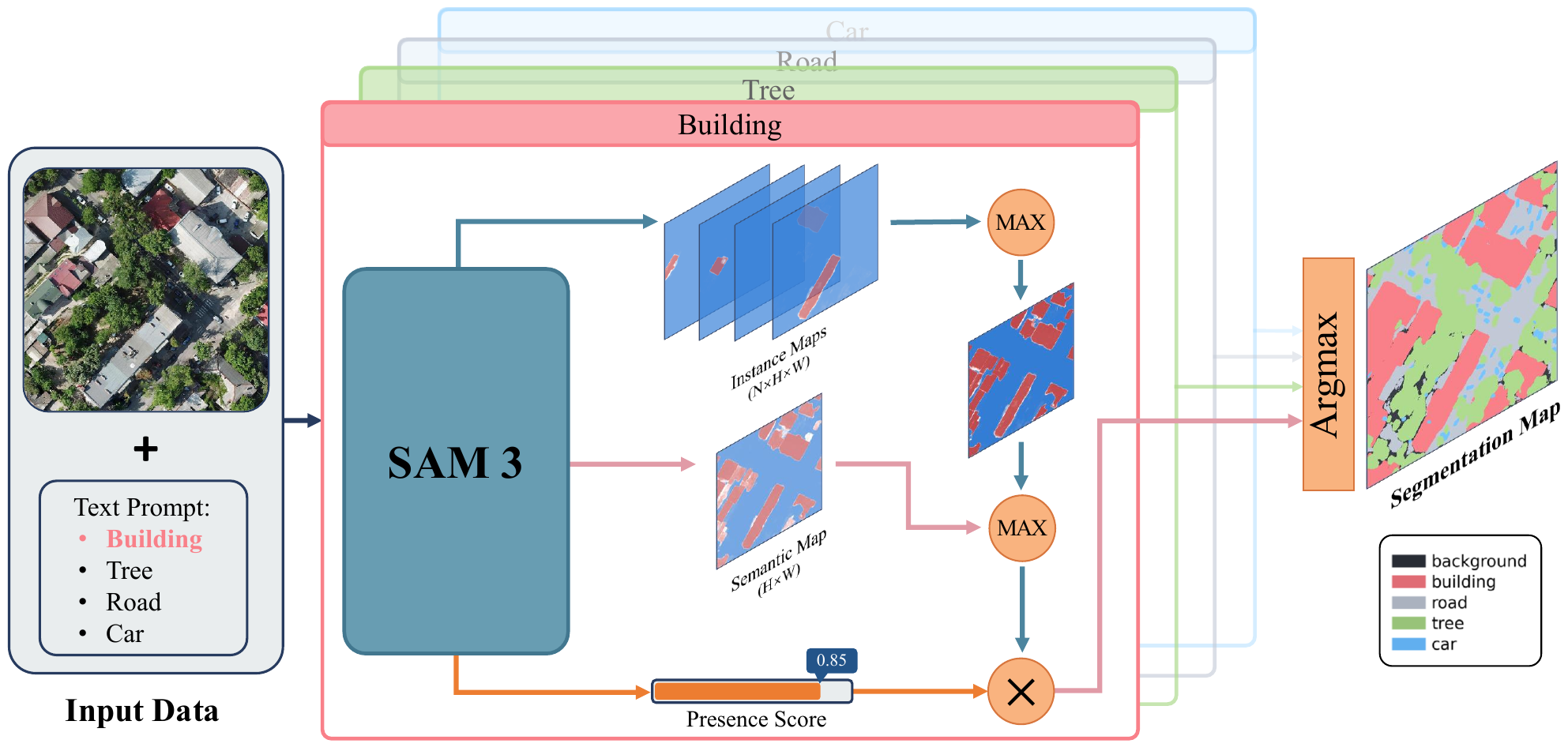}
   \caption{The overall inference pipeline of \samov. Given an input image and a list of text prompts, we leverage SAM 3's decoupled outputs. The pipeline involves: (1) instance aggregation to consolidate sparse object predictions; (2) dual-head mask fusion to combine the fine-grained instance details with the global coverage of the semantic head; and (3) presence-guided filtering (using the presence score) to suppress false positives from absent categories. \includegraphics[scale=0.04,valign=c]{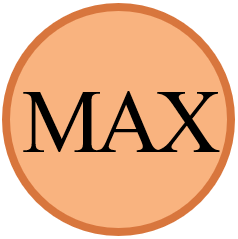} denotes the element-wise maximum operation, and \includegraphics[scale=0.026,valign=c]{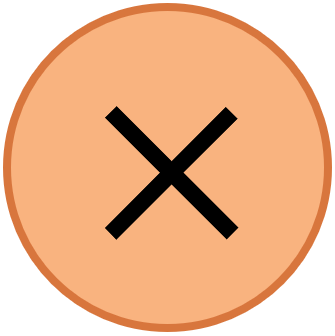} denotes multiplication.}
   \label{fig:method}
\end{figure*}

\subsection{Training-based OVSS}

Training-based OVSS methods fine-tune pre-trained VLMs on annotated datasets, typically adopting either a mask classification or a dense feature adaptation paradigm. Mask classification methods, including OpenSeg~\citep{ghiasi2022scaling}, OVSeg~\citep{liang2023open}, ZegFormer~\citep{ding2022decoupling}, and MasQCLIP~\citep{xu2023masqclip}, \textit{etc.}, leverage generated class-agnostic masks for subsequent CLIP classification. In contrast, dense feature adaptation methods, ranging from early pixel-alignment works like LSeg~\citep{li2022language} to advanced adapter-based models like SAN~\citep{xu2023side}, SED~\citep{xie2024sed} and CAT-Seg~\citep{cho2024cat}, refine CLIP's feature maps directly for dense prediction via side networks or cost aggregation. Other methods such as SegCLIP~\citep{luo2023segclip} and GroupViT~\citep{xu2022groupvit} explore weakly-supervised grouping from image-text pairs. Despite achieving strong performance, these methods require computationally expensive training on large-scale datasets, limiting their flexibility compared to training-free alternatives.

\subsection{Training-free OVSS}

To avoid training costs, training-free methods directly adapt pre-trained VLMs for dense prediction. Pure CLIP-based methods such as SCLIP~\citep{wang2023sclip} and ClearCLIP~\citep{lan2024clearclip} modify the internal mechanisms of CLIP, \textit{e.g.}, removing pooling layers or refining self-attention maps, to extract dense features~\citep{zhou2022extract, bousselham2024grounding, hajimiri2025pay, jin2025feature}. However, due to CLIP's image-level pre-training objective, these methods often suffer from coarse localization. To address this, the VFM-assisted methods~\citep{lan2025proxyclip, zhang2025corrclip, shi2025harnessing, kim2025distilling, barsellotti2024training} integrate auxiliary models like DINO~\citep{caron2021emerging, oquab2023dinov2} or SAM~\citep{kirillov2023segment, ravi2024sam} to provide structural guidance. By utilizing attention maps or object proposals from these foundation models, they achieve clearer boundaries. However, these methods operate as disjointed pipelines, relying on separate, heavy models for segmentation and recognition, which leads to significant system complexity.

\subsection{Remote Sensing OVSS}

Extending OVSS to remote sensing images faces unique challenges, such as extreme scale variations and arbitrary orientations, which often degrade the performance of methods designed for natural images. To mitigate the coarse localization of CLIP, recent methods like SegEarth-OV~\citep{li2025segearth} and GSNet~\citep{ye2025towards} employ feature upsampling modules or dual-stream architectures to incorporate domain-specific priors, while SkySense-O~\citep{zhu2025skysense} advances this by pre-training vision-centric VLMs on large-scale remote sensing data. Addressing the specific geometric complexities of aerial views, \cite{cao2024open} introduce rotation-aggregative modules to handle orientation diversity. Furthermore, AerOSeg~\citep{dutta2025aeroseg} and SCORE~\citep{huang2025score} leverage SAM features for structural guidance and regional context, while RemoteSAM~\citep{yao2025remotesam} and InstructSAM~\citep{zheng2025instructsam} utilize SAM-based pipelines to unify segmentation with broader interpretation tasks. However, these methods typically rely on complex multi-stage pipelines or require expensive domain-specific training. In contrast, we explore SAM 3~\citep{carion2025sam} as a unified, training-free framework to simplify remote sensing OVSS.

\subsection{Remote Sensing OVCD}

OVCD extends traditional closed-set remote sensing change detection to identify arbitrary change categories using textual descriptions. Early explorations leverage vision foundation models to locate generic changes; for instance, AnyChange~\citep{zheng2024segment} utilizes latent matching of SAM features for class-agnostic mask generation. Advancing beyond generic changes, SCM~\citep{tan2023segment} introduces a piecewise semantic attention mechanism combining FastSAM~\citep{zhao2023fast} and CLIP to extract building-specific changes. To systematically address arbitrary categories, DynamicEarth~\citep{li2026dynamicearth} formally defines the task and proposes decoupled frameworks like Mask-Compare-Identify (M-C-I). Concurrently, other works explore diverse strategies to bridge modality and temporal gaps: AdaptOVCD~\citep{dou2026adaptovcd} and UniVCD~\citep{zhu2025univcd} employ multi-level information fusion or feature alignment modules, while Semantic-CD~\citep{zhu2025semantic} introduces decoupled multi-task learning to refine semantic changes.
Despite this progress, most existing OVCD methods rely on disjointed multi-model ensembles, leading to system complexity. Recently, researchers have turned to the unified architecture of SAM~3 to simplify the pipeline. However, methods like CoRegOVCD~\citep{tang2026coregovcd} still rely on cumbersome post-processing steps, such as dense posterior calibration and regional consensus, to refine SAM 3's predictions. Concurrently, built upon an early version of our \samov~\citep{li2025segearthov3}, Seg2Change~\citep{su2026seg2change} designs an additional category-agnostic change adapter, and OmniOVCD~\citep{zhang2026omniovcd} attempts to streamline the process via instance decoupling. Nevertheless, their reliance on pure instance-level matching or external modules remains sensitive to registration noise in amorphous scenes. To address this, we extend our \samov framework to OVCD with a joint instance- and pixel-level verification strategy. By leveraging dense visual features and logits from SAM~3 as constraints, our simple yet effective design filters out pseudo-changes and registration errors, achieving robust open-vocabulary change detection without requiring complex pipelines or heavy post-processing.

\section{Methods}
\label{sec:methods}

\subsection{Preliminaries}

We adopt SAM 3 as our foundational architecture. Unlike standard semantic segmentors that map an image $I \in \mathbb{R}^{H \times W \times 3}$ directly to a label map $L \in \mathbb{R}^{H \times W}$, SAM 3 operates as a prompt-conditioned predictor. Given an image $I$ and a specific text prompt $t$ (\textit{e.g.}, a category name ``building''), the model predicts the probability that a pixel or region belongs to the concept defined by $t$.
Architecturally, SAM 3 consists of a vision encoder and a text encoder that extract image embeddings and text embeddings, respectively. These embeddings are processed by a Fusion Encoder, which outputs prompt-conditioned image features $F_{cond}$. Based on $F_{cond}$, the model utilizes three decoupled heads to generate predictions:

\begin{itemize}
    \item \textbf{Presence Head:} Predicts a scalar score $S_{pres} \in [0, 1]$, indicating the global probability that the concept $t$ exists in the image.
    \item \textbf{Semantic Segmentation Head:} A dense prediction module (FCN-style) that maps $F_{cond}$ to a semantic probability map $P_{sem} \in [0, 1]^{H \times W}$.
    \item \textbf{Transformer Decoder (Instance Head):} A query-based module that outputs a set of $N$ instance predictions $\{(P_{inst}^{(k)}, s_{conf}^{(k)})\}_{k=1}^N$. Here, $P_{inst}^{(k)} \in [0, 1]^{H \times W}$ is the probability map for the $k$-th object query, and $s_{conf}^{(k)} \in [0, 1]$ is its associated confidence score.
\end{itemize}

For OVSS, a naive baseline is to rely solely on the instance predictions from the Transformer decoder~\citep{carion2025sam}, aggregating them to form a segmentation mask. However, in remote sensing, this baseline often overlooks amorphous regions and is prone to false positives when querying a large vocabulary. Motivated by this, we introduce some strategies to address these limitations.

\subsection{\samov Pipeline}

We propose a training-free inference strategy designed to tackle the specific challenges of OVSS in remote sensing scenarios, as shown in Figure~\ref{fig:method}. Our pipeline processes each category in the vocabulary $\mathcal{V}$ sequentially and aggregates the results into a final segmentation map.

\noindent
\textbf{Instance Aggregation.}
Remote sensing scenes often contain dense clusters of small, identical objects (\textit{e.g.}, vehicles in a parking lot, ships in a harbor). Standard semantic segmentation methods often lead to boundary adhesion in dense clusters. To address this, we first leverage the Transformer decoder, which treats objects as discrete queries, effectively isolating individual instances even in crowded scenes. It generates a set of $N$ discrete instance predictions $\{(P_{inst}^{(k)}, s_{conf}^{(k)})\}_{k=1}^N$. We aggregate these sparse predictions into a single category-level map $P_{inst\_agg}$ by taking the maximum weighted probability at each pixel:

\begin{equation}
\begin{aligned}
  P_{inst\_agg}(h, w) = \max_{k=1}^{N} \left( P_{inst}^{(k)}(h, w) \cdot s_{conf}^{(k)} \right).
  \label{eq:instance}
\end{aligned}
\end{equation}

This aggregation effectively consolidates individual object instances into a unified semantic layer, preserving the fine-grained localization capabilities of the instance head even in crowded scenes.

\begin{figure}[t]
  \centering
   \includegraphics[width=1.0\linewidth]{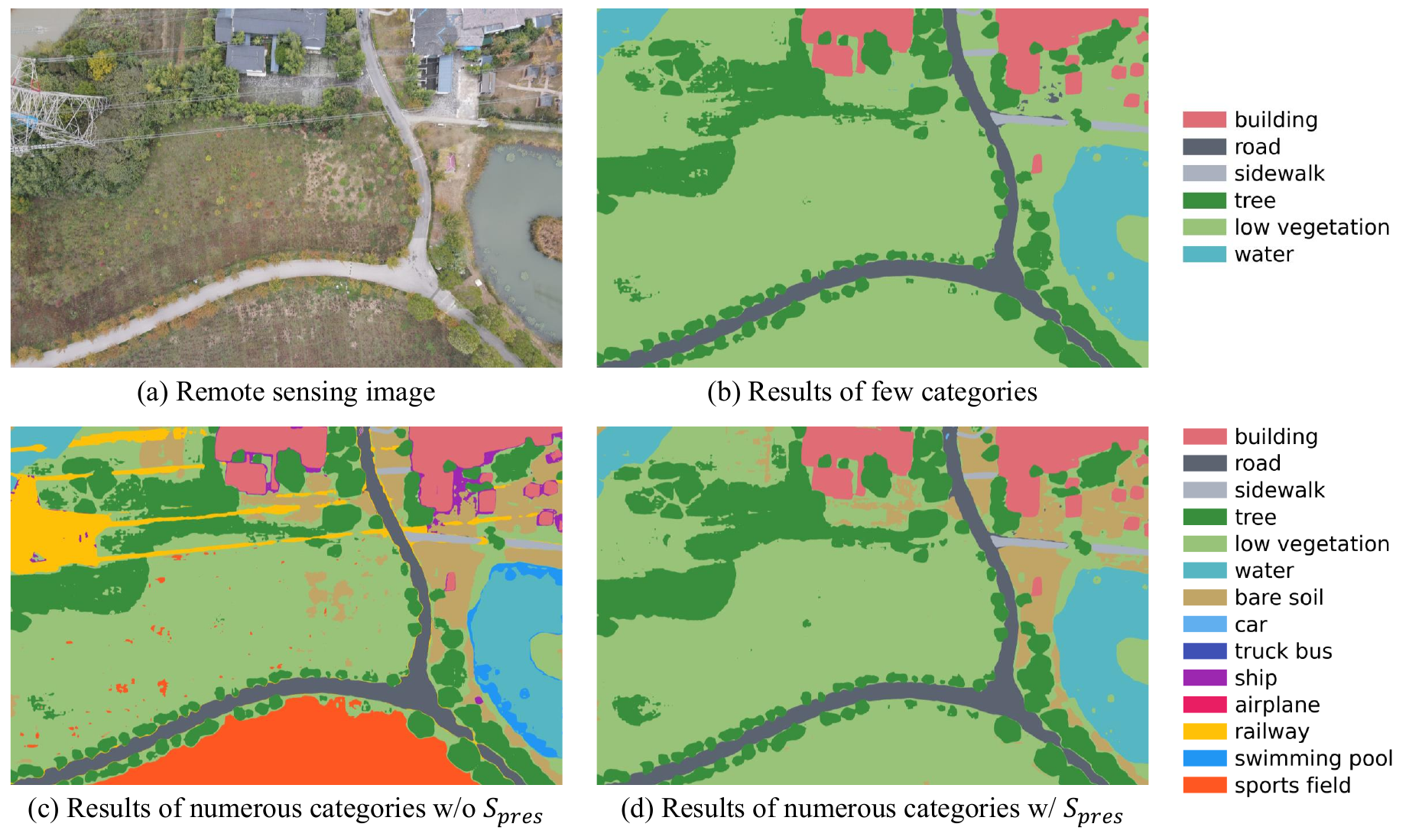}
   \caption{Impact of vocabulary size and our filtering strategy. Querying a vast vocabulary introduces severe noise due to distracting candidates (b to c). Our presence-guided filtering leverages presence scores to suppress absent categories, effectively eliminating interference and restoring segmentation quality.}
   \label{fig:presence}
   \vspace{-1em}
\end{figure}

\noindent
\textbf{Dual-Head Mask Fusion.}
While the instance head excels at delineating countable objects (\textit{i.e.}, ``things''), it may produce fragmented predictions for large-scale continuous and amorphous regions (\textit{i.e.}, ``stuff'') like road or bareland, which are prevalent in remote sensing images, as shown in Figure~\ref{fig:motivation}. Conversely, the semantic head provides dense, global coverage but often blurs the boundaries of small targets or misses them entirely. To reconcile these complementary strengths, we fuse the aggregated instance map $P_{inst\_agg}$ with the dense probability map $P_{sem}$ from the semantic head using a max-fusion strategy:

\begin{equation}
\begin{aligned}
  P_{fused}(h, w) = \max \left( P_{sem}(h, w), P_{inst\_agg}(h, w) \right).
  \label{eq:semantic}
\end{aligned}
\end{equation}

This fusion ensures robust segmentation across diverse categories, capturing both the distinct boundaries of small instances and the completeness of large-scale amorphous regions.

\noindent
\textbf{Presence-Guided Filtering.}
A critical challenge in OVSS arises from querying a massive global vocabulary against a localized image patch. Although the vocabulary $\mathcal{V}$ might enumerate a comprehensive list of global land-cover types (\textit{e.g.}, various biomes, infrastructure), a single inference patch restricts the view to a small geographical extent (\textit{e.g.}, a few hundred meters). This results in a high category sparsity, where the valid targets in a given image are only a small subset of $\mathcal{V}$. The model becomes prone to hallucinating absent categories due to textural ambiguities common in geospatial data, \textit{e.g.}, confusing ``low vegetation'' with ``sports field'', as shown in Figure~\ref{fig:presence}.

To mitigate this, we utilize the global presence score $S_{pres}$ to explicitly suppress irrelevant categories. We apply a soft gating operation, $P_{final}^{(c)} = P_{fused}^{(c)} \cdot S_{pres}^{(c)}$, which reduces the weight of the probability maps of categories predicted to be absent. Finally, we assign each pixel to the category with the highest probability:


\begin{equation}
\begin{aligned}
  M(h, w) = \arg\max_{c \in \mathcal{V}} P_{final}^{(c)}(h, w).
  \label{eq:argmax}
\end{aligned}
\end{equation}

To handle ambiguous regions, pixels with a maximum probability below a threshold $\tau$ are assigned to the ``background'' category (if it exists). Note that for categories with multiple prompts, we select the one with the highest probabilities to ensure robustness.

\begin{table*}
  \caption{Open-vocabulary semantic segmentation quantitative comparison on remote sensing datasets. Evaluation metric: mIoU. \textcolor{tabred}{\textbf{Best}} and \textcolor{tabblue}{\textbf{second best}} performances are highlighted. SCAN, SAN, SED, Cat-Seg, OVRS, GSNet, RSKT-Seg are tuned on  dataset (7,002 images with 17 categories). SkySense-O is tuned on Sky-SA dataset (35,000 images with 1,763 categories). ``Oracle'' is achieved by a fully supervised SegFormer-b0~\citep{xie2021segformer} model using full training data.}
  \label{table_main}
  \centering
  \scalebox{0.9}{
  \begin{tabular}{@{}lcccccccc|c@{}}
    \toprule[1pt]
    Methods & OpenEarthMap & LoveDA & iSAID & Potsdam & Vaihingen & UAVid$^{img}$ & UDD5 & VDD & Avg \\
    \midrule
    \rowcolor{ForestGreen!8}
    \multicolumn{10}{c}{\textbf{\emph{Training on remote sensing segmentation data}}}   \\ 
    SCAN$_{\textcolor{gray}{\text{\tiny \textit{CVPR2024}}}}$~\citep{liu2024open} & - & 23.2 & 44.3 & 27.5 & 15.2 & 20.3 & 34.1 & 29.2 & - \\
    SAN$_{\textcolor{gray}{\text{\tiny \textit{CVPR2023}}}}$~\citep{xu2023side} & - & 25.3 & 49.6 & 37.3 & 39.2 & 23.5 & 37.2 & 35.8 & - \\
    SED$_{\textcolor{gray}{\text{\tiny \textit{CVPR2024}}}}$~\citep{xie2024sed} & - & 24.6 & 51.2 & 29.4 & 39.0 & 21.3 & 35.7 & 32.5 & - \\
    Cat-Seg$_{\textcolor{gray}{\text{\tiny \textit{CVPR2024}}}}$~\citep{cho2024cat} & - & 28.6 & 53.3 & 35.8 & 42.3 & 25.7 & 40.2 & 39.1 & - \\
    OVRS$_{\textcolor{gray}{\text{\tiny\textit{TGRS2025}}}}$~\citep{cao2024open} & - & 31.5 & 52.7 & 36.4 & 43.5 & 24.1 & 40.8 & 37.2 & - \\
    GSNet$_{\textcolor{gray}{\text{\tiny\textit{AAAI2025}}}}$~\citep{ye2025towards} & - & 32.5 & \textcolor{tabblue}{\textbf{53.7}} & 37.9 & 44.1 & 24.2 & 40.9 & 37.3 & - \\
    RSKT-Seg$_{\textcolor{gray}{\text{\tiny\textit{AAAI2026}}}}$~\citep{li2025exploring} & - & 33.2 & \textcolor{tabred}{\textbf{54.3}} & 38.4 & 42.7 & 25.7 & 42.1 & 39.7 & - \\
    SkySense-O$_{\textcolor{gray}{\text{\tiny\textit{CVPR2025}}}}$~\citep{zhu2025skysense} & \textcolor{tabblue}{\textbf{40.8}} & \textcolor{tabblue}{\textbf{38.3}} & 43.9 & \textcolor{tabblue}{\textbf{54.1}} & \textcolor{tabblue}{\textbf{51.6}} & - & - & - & - \\
    \midrule
    \rowcolor{ForestGreen!8}
    \multicolumn{10}{c}{\textbf{\emph{Training-free}}}   \\ 
    CLIP$_{\textcolor{gray}{\text{\tiny \textit{ICML2021}}}}$~\citep{radford2021learning} & 12.0 & 12.4 & 7.5 & 15.6 & 10.8 & 10.9 & 9.5 & 14.2 & 11.4 \\
    MaskCLIP$_{\textcolor{gray}{\text{\tiny \textit{ECCV2022}}}}~\citep{zhou2022extract}$ & 25.1 & 27.8 & 14.5 & 33.9 & 29.9 & 28.6 & 32.4 & 32.9 & 27.2 \\
    SCLIP$_{\textcolor{gray}{\text{\tiny \textit{ECCV2024}}}}$~\citep{wang2023sclip} & 29.3 & 30.4 & 16.1 & 39.6 & 35.9 & 31.4 & 38.7 & 37.9 & 31.1 \\
    GEM$_{\textcolor{gray}{\text{\tiny \textit{CVPR2024}}}}$~\citep{bousselham2024grounding} & 33.9 & 31.6 & 17.7 & 39.1 & 36.4 & 33.4 & 41.2 & 39.5 & 32.3 \\
    ClearCLIP$_{\textcolor{gray}{\text{\tiny \textit{ECCV2024}}}}$~\citep{lan2024clearclip} & 31.0 & 32.4 & 18.2 & 42.0 & 36.2 & 36.2 & 41.8 & 39.3 & 33.4 \\
    SegEarth-OV$_{\textcolor{gray}{\text{\tiny \textit{CVPR2025}}}}$~\citep{li2025segearth} & 40.3 & 36.9 & 21.7 & 48.5 & 40.0 & \textcolor{tabblue}{\textbf{42.5}} & \textcolor{tabblue}{\textbf{50.6}} & 45.3 & 39.2 \\
    ProxyCLIP$_{\textcolor{gray}{\text{\tiny \textit{ECCV2024}}}}$~\citep{lan2025proxyclip} & 38.9 & 34.3 & 21.8 & 49.0 & 47.5 & 35.8 & 40.8 & \textcolor{tabblue}{\textbf{47.8}} & 39.5 \\
    CorrCLIP$_{\textcolor{gray}{\text{\tiny \textit{ICCV2025}}}}$~\citep{zhang2025corrclip} & 32.9 & 36.9 & 25.5 & 51.9 & 47.0 & 38.3 & 46.1 & 47.3 & \textcolor{tabblue}{\textbf{40.7}} \\
    \samov & \textcolor{tabred}{\textbf{42.9}} & \textcolor{tabred}{\textbf{47.4}} & 27.6 & \textcolor{tabred}{\textbf{57.8}} & \textcolor{tabred}{\textbf{60.8}} & \textcolor{tabred}{\textbf{54.7}} & \textcolor{tabred}{\textbf{71.7}} & \textcolor{tabred}{\textbf{64.5}} & \textcolor{tabred}{\textbf{53.4}} \\
    \midrule
    \textcolor{tablegray}{Oracle} & \textcolor{tablegray}{64.4} & \textcolor{tablegray}{50.0} & \textcolor{tablegray}{36.2} & \textcolor{tablegray}{74.3} & \textcolor{tablegray}{61.2} & \textcolor{tablegray}{59.7} & \textcolor{tablegray}{56.5} & \textcolor{tablegray}{62.9} & \textcolor{tablegray}{58.2} \\
    \bottomrule
  \end{tabular}}
\end{table*}

\subsection{Extension to OVCD}

Building on the above, we extend \samov to OVCD. In zero-shot scenarios, comparing bi-temporal segmentation results directly is highly susceptible to pseudo-changes caused by inconsistent VLM recognitions and spatial misalignments. To mitigate this, we propose a joint instance- and pixel-level verification strategy.

\noindent
\textbf{Pixel-level Comparison.}
To suppress pseudo-changes caused by appearance variations (\textit{e.g.}, illumination or seasonal changes), we utilize the dense visual features from the SAM 3 image encoder as a structural constraint. Let $I_1$ and $I_2$ be the bi-temporal images, and $F_1, F_2$ be their corresponding extracted visual features by the the vision encoder of SAM~3. We compute the cosine similarity between $F_1$ and $F_2$, normalizing the similarity scores to a range of $[0, 1]$. Then a feature change map $F_{change}$ is derived by subtracting this similarity from 1, where regions with high similarity yield values close to 0.

Next, we construct a joint energy map $E^{(c)}$ to identify significant semantic change for a specific category $c$. For a given category, let $P_1^{(c)}$ and $P_2^{(c)}$ denote the probability maps generated by \samov. The energy map is formulated as:
\begin{equation}
    E^{(c)} = \max(P_1^{(c)}, P_2^{(c)}) \cdot |P_1^{(c)} - P_2^{(c)}| \cdot F_{change}.
\end{equation}
The term $\max(P_1^{(c)}, P_2^{(c)})$ ensures that the pixel is confidently recognized as category $c$ in at least one of the timestamps, while $|P_1^{(c)} - P_2^{(c)}|$ quantifies the magnitude of the semantic change for that category. The $F_{change}$ represents category-agnostic visual differences, to encompass regions that are not activated by text prompts. Finally, we apply the OTSU~\citep{otsu1975threshold} thresholding to $E^{(c)}$ to generate a binary pixel-level change mask $D_{pixel}^{(c)}$.

\noindent
\textbf{Instance-level Comparison.}
Remote sensing multi-temporal images frequently suffer from spatial mis-alignments, causing unchanged objects to produce false positive. To overcome this, following~\citep{zhang2026omniovcd}, we convert the continuous probability maps of \samov into discrete instance-level predictions. To tolerate registration drifts, we apply morphological dilation to the discrete masks. Specifically, an instance in one temporal image is flagged as changed only if its coverage ratio by the dilated mask of the other image falls below a predefined threshold. This instance-level comparison is performed symmetrically for both temporal directions to yield an instance-level change mask $D_{inst}^{(c)}$.

To ensure the highest reliability and minimize false alarms induced by VLM recognition inconsistencies, a pixel is classified as a valid change only if it is confirmed by comparisons at both the pixel-level and the instance-level, \textit{i.e.}, $D_{inst}^{(c)} \cap D_{pixel}^{(c)}$. By employing this joint verification, our method effectively filters out noise while maintaining sensitivity to true temporal changes for arbitrary categories.

\section{Experiments}
\label{sec:exps}

\begin{figure*}[t]
  \centering
   \includegraphics[width=1.0\linewidth]{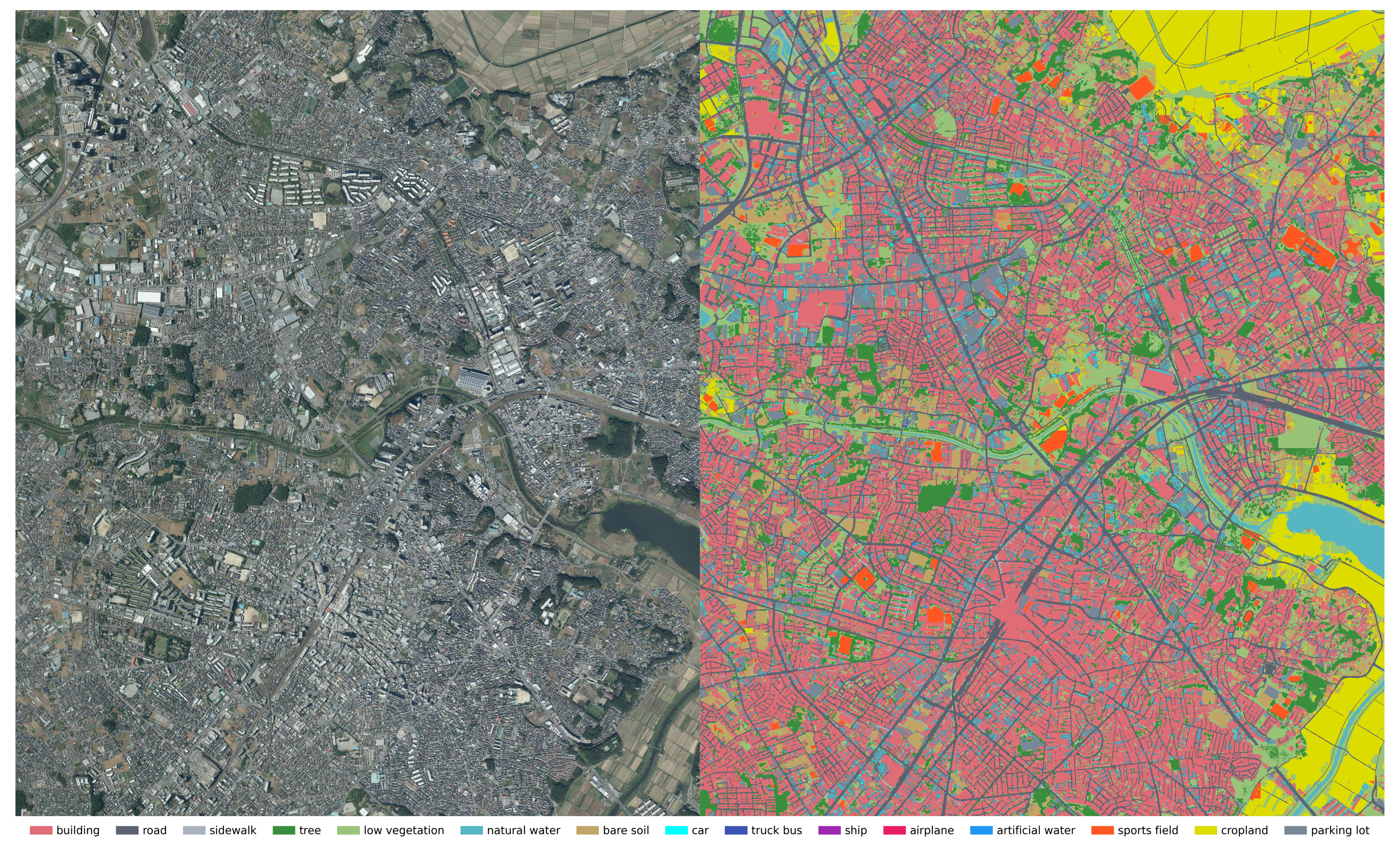}
   \caption{Inference results of \samov on a remote sensing image exceeding 10k$\times$10k resolution. The image originates from~\citep{chen2023land}.}
   \label{fig:vis}
\end{figure*}

\begin{table*}[t]
  \caption{Open-vocabulary building / road / flood extraction quantitative comparison on remote sensing datasets. Evaluation metric: IoU of the foreground class, \textit{i.e.} building, road or flood. \textcolor{tabred}{\textbf{Best}} and \textcolor{tabblue}{\textbf{second best}} performances are highlighted.}
  \label{table_main_br}
  \centering
  \scalebox{0.8}{
  \begin{tabular}{@{}lcccc|cccc|c@{}}
    \toprule
    \multirow{2}{*}{Method} & \multicolumn{4}{c|}{\color{red!50!black}Building Extraction} & \multicolumn{4}{c|}{\color{yellow!50!black}Road Extraction} & \multicolumn{1}{c}{\color{blue!50!black}Flood Detection}\\
    & WHU$^{Aerial}$ & WHU$^{Sat.\mathrm{II}}$  & Inria & xBD$^{pre}$ & CHN6-CUG & DeepGlobe & Massachusetts & SpaceNet & WBS-SI \\
    \midrule
    CLIP~\citep{radford2021learning} & 17.7 & 3.5 & 19.6 & 16.0 & 7.7 & 3.9 & 4.9 & 7.1 & 18.6 \\
    MaskCLIP~\citep{zhou2022extract} & 29.8 & 14.0 & 33.4 & 29.2 & 28.1 & 13.2 & 10.6 & 20.8 & 39.8 \\
    SCLIP~\citep{wang2023sclip} & 33.4 & 21.0 & 34.9 & 25.9 & 21.1 & 7.0 & 7.4 & 14.9 & 32.1 \\
    GEM~\citep{bousselham2024grounding} & 24.4 & 13.6 & 28.5 & 20.8 & 13.4 & 4.7 & 5.1 & 11.9 & 39.5 \\
    ClearCLIP~\citep{lan2024clearclip} & 36.6 & 20.8 & 39.0 & 30.1 & 25.5 & 5.7 & 6.4 & 16.3 & 44.9\\
    SegEarth-OV~\citep{li2025segearth} & \textcolor{tabblue}{\textbf{49.2}} & \textcolor{tabblue}{\textbf{28.4}} & \textcolor{tabblue}{\textbf{44.6}} & \textcolor{tabblue}{\textbf{37.0}} & \textcolor{tabblue}{\textbf{35.4}} & \textcolor{tabblue}{\textbf{17.8}} & \textcolor{tabblue}{\textbf{11.5}} & \textcolor{tabblue}{\textbf{23.8}} & \textcolor{tabblue}{\textbf{60.2}} \\
    \midrule
    \samov & \textcolor{tabred}{\textbf{86.9}} & \textcolor{tabred}{\textbf{44.2}} & \textcolor{tabred}{\textbf{72.4}} & \textcolor{tabred}{\textbf{64.3}} & \textcolor{tabred}{\textbf{49.6}} & \textcolor{tabred}{\textbf{39.3}} & \textcolor{tabred}{\textbf{27.7}} & \textcolor{tabred}{\textbf{35.6}} & \textcolor{tabred}{\textbf{75.6}} \\
    \bottomrule
  \end{tabular}}
  \vspace{-1em}
\end{table*}

\subsection{Setup}

\textbf{Datasets.} We evaluate the performance of \samov across three main open-vocabulary tasks, involving a total of 24 remote sensing benchmarks and 3 general image benchmarks to ensure a comprehensive assessment.

\begin{itemize}
    \item 2D Remote Sensing Semantic Segmentation: We use 20 datasets in total. This includes eight multi-class remote sensing benchmarks (OpenEarthMap~\citep{xia2023openearthmap}, LoveDA~\citep{loveda}, iSAID~\citep{waqas2019isaid}, Potsdam, Vaihingen\footnote{https://www.isprs.org/education/benchmarks/UrbanSemLab}, UAVid~\citep{LYU2020108}, UDD5~\citep{chen2018large} and VDD~\citep{cai2023vdd}) and nine binary extraction datasets focusing on critical geospatial targets (buildings: WHU$^{Aerial}$~\citep{ji2018fully}, WHU$^{Sat.\mathrm{II}}$~\citep{ji2018fully}, Inria~\citep{maggiori2017can}, and xBD~\citep{gupta2019xbddatasetassessingbuilding}; roads: CHN6-CUG~\citep{zhu2021global}, DeepGlobe~\citep{demir2018deepglobe}, Massachusetts~\citep{MnihThesis}, and SpaceNet~\citep{van2018spacenet}; and floods: WBS-SI\footnote{https://www.kaggle.com/datasets/shirshmall/water-body-segmentation-in-satellite-images}). Specifically, we evaluate the performance of the OVSS method on three GF-series datasets, including the GID dataset (GF-2)~\citep{tong2020land}, the GF-7 building dataset\footnote{https://figshare.com/articles/dataset/GF\_LowGradeRoadDataset \_zip/24457369/2?file=42966904}, and the Low-Grade Road Dataset (GF-2)~\citep{chen2023leveraging}.

    \item Remote Sensing Change Detection: We evaluate our method on three representative change detection benchmarks: LEVIR-CD~\citep{chen2020spatial}, WHU-CD~\citep{ji2018fully}, and S2Looking~\citep{shen2021s2looking}.

    \item 3D Remote Sensing Semantic Segmentation: We utilize the STPLS3D~\citep{chen2022stpls3d} dataset, performing multi-view projection from 2D predictions to 3D point clouds for evaluation.

    \item 2D general image segmentation: To demonstrate universality of \samov, we also include three general scene benchmarks: Pascal VOC20~\citep{everingham2010pascal}, COCO Stuff~\citep{caesar2018coco}, and Cityscapes~\citep{cordts2016cityscapes}.

\end{itemize}

The evaluation metric is mIoU for multi-class segmentation and IoU of the foreground class for binary segmentation tasks.


\noindent
\textbf{Implementation Details.} We use the official SAM 3 model equipped with the Perception Encoder-Large+ (PE-L+)~\citep{bolya2025perception} backbone. Input images are resized to 1008$\times$1008. Text prompts are derived directly from category names (\textit{e.g.}, ``building'', ``road''), and some categories are augmented with synonyms. The background threshold $\tau$ and the initial confidence threshold for the Transformer decoder are slightly  adjusted for each dataset to achieve roughly optimal performance. No test-time augmentation or extra post-processing method~\citep{araslanov2020single} is used.

\begin{table}
  \caption{Comparison of different OVSS methods on GF-series datasets (GID, GF-7 Building dataset, Low-Grade Road dataset).}
  \label{}
  \centering
  \scalebox{0.95}{
  \begin{tabular}{lccc}
    \toprule
    Method & GID & GF-7 Building &  Low-Grade Road \\
    \midrule
    CLIP        & 22.7 & 19.1 & 3.3  \\
    MaskCLIP    & 35.4 & 26.1 & 9.5  \\
    GEM         & 40.1 & 23.3 & 4.9  \\
    SCLIP       & 37.6 & 27.2 & 5.7  \\
    ClearCLIP   & 39.7 & 28.2 & 5.1  \\
    SegEarth-OV & \textbf{46.3} & 32.1 & 9.4  \\
    ProxyCLIP   & 42.8 & 34.9 & 22.1 \\
    CorrCLIP    & 40.9 & 32.2 & 10.5 \\
    \samov         & 42.2 & \textbf{58.7} & \textbf{60.2} \\
    \bottomrule
  \end{tabular}}
\end{table}

\begin{table}[t]
  \centering
  \caption{Ablation study on dual-head mask fusion. ``Instance Only'' denotes using predictions solely from the Transformer decoder, while ``Semantic Only'' relies exclusively on the semantic segmentation head. \textbf{Bold} indicates the best performance.}
  \resizebox{0.47\textwidth}{!}{
  \begin{tabular}{lcccc}
    \toprule
    Method & LoveDA & Uavid & {\color{red!50!black}xBD$^{pre}$} & {\color{yellow!50!black}CHN6-CUG} \\
    \midrule
    Instance Only & 32.2 & 50.4 & 61.4 & 38.4 \\
    Semantic Only & 35.4 & 47.1 & 44.9 & 39.5 \\
    \samov & \textbf{47.4} & \textbf{54.7} & \textbf{64.3} & \textbf{49.6} \\
    \bottomrule
  \end{tabular}}
  \label{tab:ab}
\end{table}

\begin{table*}[t]
\caption{Quantitative evaluation of different methods on the change detection dataset. $\dagger$ denotes results from~\citep{su2026seg2change}. The {\color{CarnationPink}{pink}} background indicates methods based on \samov. JIPS denotes our joint instance- and pixel-level verification strategy. JIPS denotes our joint instance- and pixel-level verification strategy.}
\label{tab:ovcd}
\centering
\scalebox{0.99}{
\begin{tabular}{l|c|l|cc|cc|cc}
\toprule
\multicolumn{1}{c|}{\multirow{2}{*}{Method}} & \multirow{2}{*}{Base Model} & \multicolumn{1}{c|}{\multirow{2}{*}{Comparator}} & \multicolumn{2}{c|}{WHU-CD} & \multicolumn{2}{c|}{LEVIR-CD} & \multicolumn{2}{c}{S2Looking} \\ \cmidrule{4-9} 
\multicolumn{1}{c|}{} &  & \multicolumn{1}{c|}{} & $F1^c$ & $IoU^c$ & $F1^c$ & $IoU^c$ & $F1^c$ & $IoU^c$ \\ \midrule
PCA-KMeans~\citep{celik2009unsupervised} & / & KMeans & 14.3 & 7.7 & 10.0 & 5.2 & - & - \\
CVA~\citep{bovolo2006theoretical} & / & CVA Match & 7.2 & 3.7 & 9.3 & 4.9 & 5.8 & - \\
DCVA~\citep{saha2019unsupervised} & / & CVA Match & 20.6 & 11.5 & 13.9 & 7.4 & - & - \\
UCD-SCM~\citep{tan2023segment} & FastSAM + CLIP & OTSU & 32.1 & 19.1 & 32.4 & 19.3 & - & - \\
AnyChange~\citep{zheng2024segment} & SAM & Latent Match & 28.1 & 16.4 & 32.7 & 19.5 & 6.4 & - \\
\rowcolor{CarnationPink!15} AnyChange$\dagger$~\citep{zheng2024segment} & \samov & Latent Match & 69.3 & 53.0 & 72.3 & 56.6 & - & - \\
Inst-CEG~\citep{li2024semicd} & APE~\citep{shen2024aligning} & CEG & 62.5 & 45.5 & 63.3 & 46.3 & - & - \\
\rowcolor{CarnationPink!15} Inst-CEG$\dagger$~\citep{li2024semicd} & \samov & CEG & 71.4 & 55.5 & 70.6 & 54.6 & - & - \\
DynamicEarth~\citep{li2026dynamicearth} & SAM2 + SegEarth-OV & DINOv2 & 57.4 & 40.2 & 50.5 & 33.8 & 37.6 & 23.1 \\
DynamicEarth~\citep{li2026dynamicearth} & APE~\citep{shen2024aligning} & DINOv2 & 75.9 & 61.1 & 66.7 & 50.0 & 21.1 & 11.8 \\
\rowcolor{CarnationPink!15} DynamicEarth$\dagger$~\citep{li2026dynamicearth} & \samov & DINOv2 & 79.7 & 66.2 & 72.0 & 56.2 & - & - \\
\rowcolor{CarnationPink!15} OmniOVCD~\citep{zhang2026omniovcd} & \samov & SFID & 79.9 & 66.5 & 80.4 & 67.2 & 39.4 & 24.5 \\
\rowcolor{CarnationPink!15} Seg2Change~\citep{su2026seg2change} & \samov & CACH & 86.2 & 75.7 & 78.7 & 64.9 & - & - \\ 
\midrule
\samov-CD (ours) & \samov & JIPS (ours) & \textbf{88.0} & \textbf{78.6} & \textbf{85.3} & \textbf{74.3} & \textbf{48.8} & \textbf{32.3} \\
\bottomrule
\end{tabular}
}
\end{table*}

\begin{table*}[t]
\caption{Quantitative evaluation of different methods on the STPLS3D-WMSC dataset.}
\label{tab:3d_evaluation}
\centering
\scalebox{0.9}{
\begin{tabular}{llccccccc}
\toprule
\multirow{2}{*}{Training sets} & \multirow{2}{*}{Methods} & \multirow{2}{*}{mIoU (\%)} & \multicolumn{6}{c}{Per Class IoU (\%)} \\ \cmidrule{4-9} 
 &  &  & Ground & Building & Tree & Car & Light pole & Fence \\ \midrule
\multirow{5}{*}{Real 3D Data} & PointTransformer~\citep{zhao2021point} & 36.27 & 39.95 & 20.88 & 62.57 & 36.13 & 49.32 & 8.76 \\
 & RandLA-Net~\citep{hu2020randla} & 42.33 & 46.13 & 24.23 & 72.46 & 53.37 & 44.82 & 12.95 \\
 & SCF-Net~\citep{fan2021scf} & 45.93 & 68.77 & 37.27 & 65.49 & 51.50 & 31.22 & \textbf{21.34} \\
 & MinkowskiNet~\citep{choy20194d} & 46.52 & 64.22 & 29.95 & 61.33 & 45.96 & 65.25 & 12.43 \\
 & KPConv~\citep{thomas2019kpconv} & 45.22 & 60.87 & 32.13 & 69.05 & \textbf{53.80} & 52.08 & 3.40 \\ \midrule
\multirow{5}{*}{\makecell[l]{Synthetic\\3D Data}} & PointTransformer~\citep{zhao2021point} & 45.73 & 84.12 & 73.37 & 60.60 & 16.96 & 27.23 & 12.10 \\
 & RandLA-Net~\citep{hu2020randla} & 45.03 & 76.78 & 57.74 & 56.08 & 28.44 & 40.36 & 10.78 \\
 & SCF-Net~\citep{fan2021scf} & 47.82 & 77.51 & 68.68 & 56.81 & 29.87 & 42.53 & 11.52 \\
 & MinkowskiNet~\citep{choy20194d} & 50.78 & 85.23 & 72.66 & 64.80 & 31.31 & 36.85 & 13.83 \\
 & KPConv~\citep{thomas2019kpconv} & 49.16 & 85.50 & 70.65 & 63.84 & 28.75 & 32.97 & 13.22 \\ \midrule
\multirow{5}{*}{\makecell[l]{Real+Synthetic\\3D Data}} & PointTransformer~\citep{zhao2021point} & 47.64 & 80.19 & 76.35 & 57.13 & 36.35 & 23.72 & 12.10 \\
 & RandLA-Net~\citep{hu2020randla} & 50.53 & 82.90 & 66.59 & 63.77 & 33.91 & 41.84 & 14.19 \\
 & SCF-Net~\citep{fan2021scf} & 50.65 & 77.80 & 58.98 & 64.86 & 46.37 & 40.50 & 15.41 \\
 & MinkowskiNet~\citep{choy20194d} & 51.35 & 80.86 & 74.03 & 59.21 & 31.72 & 45.51 & 16.79 \\
 & KPConv~\citep{thomas2019kpconv} & 53.73 & 87.40 & \textbf{78.51} & 66.18 & 39.63 & 41.30 & 9.34 \\ 
 \midrule
 None & \samov & \textbf{61.26} & \textbf{88.92} & 75.04 & \textbf{77.40} & 44.57 & \textbf{69.02} & 12.63 \\
 \bottomrule
\end{tabular}
}
\end{table*}

\subsection{Main Results}

\noindent
\textbf{Semantic segmentation.}
We report the quantitative comparison on eight remote sensing semantic segmentation benchmarks in Table~\ref{table_main}. Our method, \samov, achieves a new state-of-the-art, demonstrating a substantial performance leap over existing methods. Specifically, it achieves an average mIoU of 53.4\%, surpassing the best previous training-free method, CorrCLIP (40.7\% mIoU), by a remarkable margin of +12.7\% mIoU. This performance also consistently outperformed training-based OVSS methods fine-tuned on remote sensing data, \textit{e.g.}, RSKT-Seg (39.7\% mIoU) and Cat-Seg (39.1\% mIoU). A notable observation is that our zero-shot method outperforms the fully supervised Oracle on some datasets. On UDD5 and VDD datasets, \samov achieves 71.7\% mIoU and 64.5\% mIoU respectively, exceeding the Oracle baselines (56.5\% and 62.9\% mIoU). We attribute this to the fact that fully supervised models are prone to overfitting when trained on domain-specific datasets with limited scale or variations. In contrast, this surprising result suggests that the rich semantic knowledge and robust segmentation capabilities inherent in the SAM 3 foundation model can provide superior generalization and robustness without requiring any target-domain supervision. In Figure~\ref{fig:vis}, we visualize the inference results on a large-scale remote sensing image exceeding $10,000 \times 10,000$ pixels.
Crucially, this seamless large-scale segmentation prediction highlights the effectiveness of our presence-guided filtering in practical deployments. When performing sliding-window inference, querying a massive global vocabulary against small localized patches typically causes severe categorical hallucinations. By dynamically suppressing absent categories patch-by-patch, our filtering strategy effectively eliminates localized false positives, verifying its practical applicability in actual large-scale scenes.

\noindent
\textbf{Single-class extraction.}
We further evaluate the single-class land-cover extraction capability of our method on nine benchmarks focusing on buildings, roads, and floods (Table~\ref{table_main_br}). In building extraction, our method achieves unprecedented gains, reaching 86.9\% IoU on WHU$^{Aerial}$ and 72.4\% IoU on Inria, outperforming the previous state-of-the-art SegEarth-OV by massive margins of +37.7\% and +27.8\% respectively. Similarly, for road extraction, it consistently surpasses previous methods, achieving 49.6\% IoU on CHN6-CUG. In the flood detection task, \samov attains 75.6\% IoU, marking a +15.4\% improvement. These consistent and dramatic improvements across diverse geospatial targets, emphasize the robustness of \samov and the generalization capability of SAM 3.


\begin{figure}[htbp]
    \centering
    
    \begin{subfigure}[b]{0.45\textwidth}
        \centering
        \includegraphics[width=\textwidth]{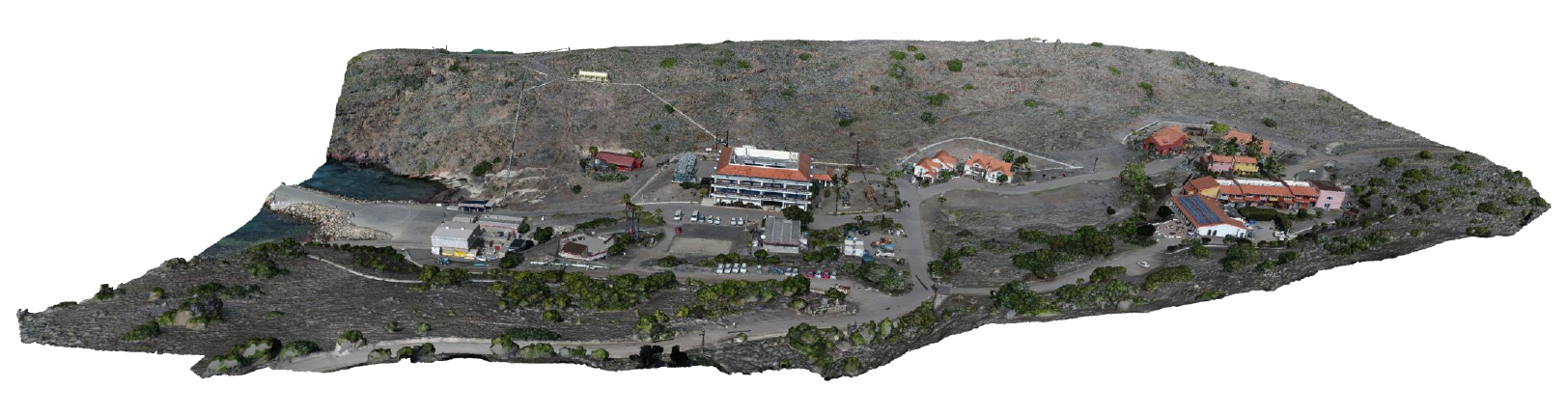}
        \caption{3D UAV scene}
        \label{fig:uav_scene}
    \end{subfigure}
    
    
    \begin{subfigure}[b]{0.45\textwidth}
        \centering
        \includegraphics[width=\textwidth]{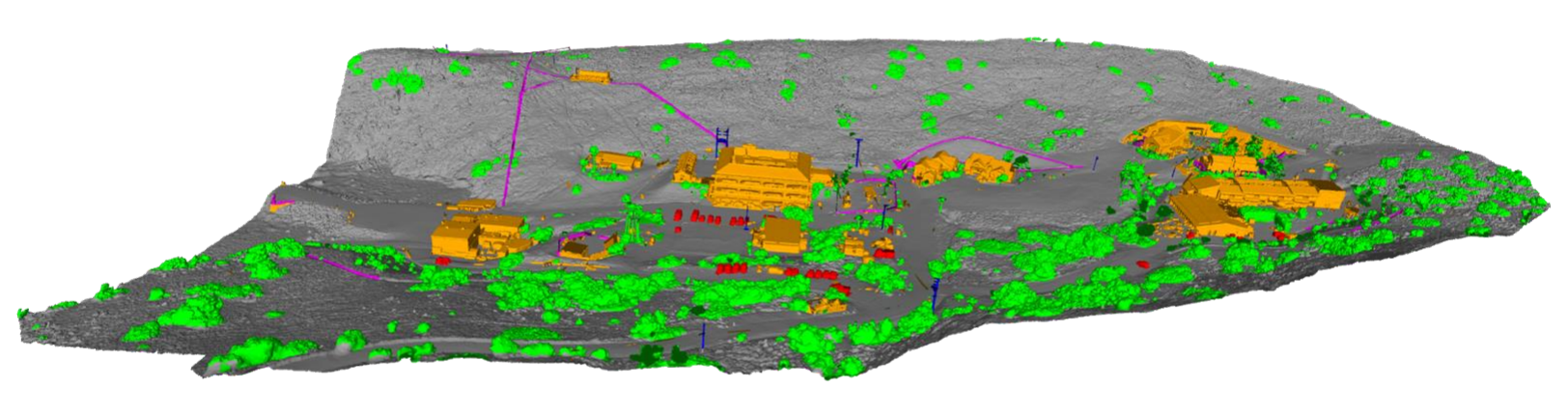}
        \caption{Ground truth}
    \end{subfigure}
    

    \begin{subfigure}[b]{0.45\textwidth}
        \centering
        \includegraphics[width=\textwidth]{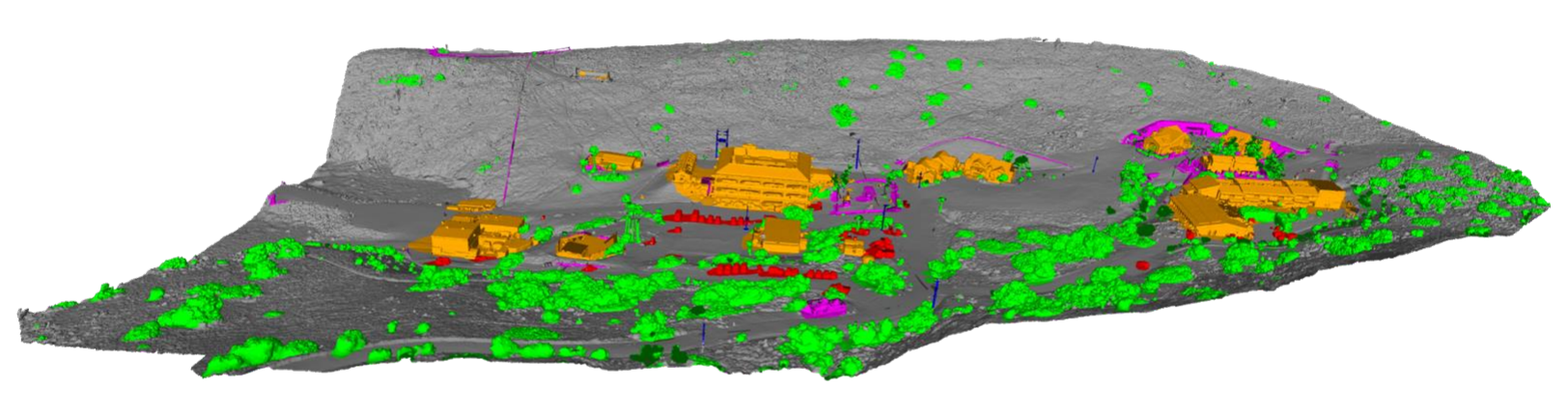}
        \caption{Our prediction}
    \end{subfigure}
    
    \caption{Visualization of our method in 3D UAV scene.}
    \label{fig:3d}
\end{figure}

\begin{table}[t]
  \centering
  \caption{Comparison with state-of-the-art OVSS methods on Pascal VOC20, COCO Stuff, and Cityscapes benchmarks. \textbf{Bold} indicates the best performance.} 
  \resizebox{0.49\textwidth}{!}{
  \begin{tabular}{lc|ccc}
    \toprule
    Method & Size & VOC20 & Stuff & City \\
    \midrule
    \rowcolor{ForestGreen!8}
    \multicolumn{5}{c}{\textbf{\emph{Training-based}}}  \\
    TCL~\citep{cha2023learning} & \multirow{5}{*}{ViT-B/16} & 83.2 & 22.4 & 24.0 \\
    CLIP-DINOiser~\citep{wysoczanska2024clip} &  & 80.9 & 24.6 & 31.7 \\
    CoDe~\citep{wu2024image}&  & - & 23.9 & 28.9 \\
    CAT-Seg~\citep{cho2024cat} &  & 94.6 & - & - \\
    
    \midrule
    \rowcolor{ForestGreen!8}
    \multicolumn{5}{c}{\textbf{\emph{Training-free}}}   \\
    CLIP~~=\cite{radford2021learning}& \multirow{14}{*}{ViT-B/16} & 41.9 & 4.4 & 5.0 \\
    MaskCLIP~\citep{zhou2022extract}& & 74.9 & 16.4 & 12.6 \\
    ClearCLIP~\citep{lan2024clearclip}& & 80.9 & 23.9 & 30.0 \\
    SCLIP~\citep{wang2023sclip}& & 80.4 & 22.4 & 32.2 \\
    ProxyCLIP~\citep{lan2025proxyclip}&  & 80.3 & 26.5 & 38.1 \\
    LaVG~\citep{lavg}&& 82.5 & 23.2 & 26.2 \\
    CLIPtrase~\citep{shao2024explore}& & 81.2 & 24.1 & - \\
    NACLIP~\citep{hajimiri2025pay} & & 83.0 & 25.7 & 38.3 \\
    Trident~\citep{shi2025harnessing} & & 84.5 & 28.3 & 42.9 \\
    ResCLIP~\citep{yang2025resclip}&& 86.0 & 24.7 & 35.9 \\
    SC-CLIP~\citep{bai2024self}&& 84.3 & 26.6 & 41.0 \\
    CLIPer~\citep{sun2025cliper}&& 85.2 & 27.5 & - \\
    CASS~\citep{kim2025distilling}&& 87.8 & 26.7 & 39.4 \\
    \rowcolor{gray!8}
    CorrCLIP~\citep{zhang2025corrclip}& & 88.8 & 31.6 & 49.4 \\
    
    \midrule
    FreeDA~\citep{barsellotti2024training}& \multirow{7}{*}{ViT-L/14} & 87.9 & 28.8 & 36.7 \\
    CaR~\citep{sun2024clip}&  & 91.4 & - & - \\
    ProxyCLIP~\citep{lan2025proxyclip}&  & 83.2 & 25.6 & 40.1 \\
    ResCLIP~\citep{yang2025resclip}&& 85.5 & 23.4 & 33.7 \\
    SC-CLIP~\citep{bai2024self}&& 88.3 & 26.9 & 41.3 \\
    CLIPer~\citep{sun2025cliper}&& 90.0 & 28.7 & - \\
    \rowcolor{gray!8}
    CorrCLIP~\citep{zhang2025corrclip}&  & 91.5 & 34.0 & 51.1 \\
    
    \midrule
    ProxyCLIP~\citep{lan2025proxyclip}& \multirow{3}{*}{ViT-H/14} & 83.3 & 26.8 & 42.0 \\
    Trident~\citep{shi2025harnessing} & & 88.7 & 28.6 & 47.6 \\
    \rowcolor{gray!8}
    CorrCLIP~\citep{zhang2025corrclip}& & 91.8 & 32.7 & 49.9 \\
    \midrule
    \samov & PE-L+/14 & \textbf{96.8} & \textbf{42.8} & \textbf{69.7} \\
    \bottomrule
  \end{tabular}
  }
  \label{tab:general}
\end{table}

\subsection{Ablation Studies}

We investigate the contribution of the dual-head fusion strategy by evaluating the performance of each head individually on four representative datasets, as listed in Table~\ref{tab:ab}. Using only Transformer decoder predictions yields strong results on object-centric tasks like building extraction, but underperforms on complex scenes, due to struggles with amorphous regions. Conversely, the ``Semantic Only'' baseline struggles with precise object boundaries, achieving only 44.9\% IoU on xBD. By fusing both heads, \samov achieves significant improvements. On LoveDA and CHN6-CUG, \samov reaches 47.4\% mIoU and 49.6\% IoU, surpassing the best single-head baselines by +12.0\% and +10.1\%. Even on the instance-heavy xBD dataset, fusion further boosts performance by 2.9\%. These results validate our hypothesis that the semantic and instance heads provide complementary information, \textit{i.e.}, global coverage and fine-grained precision, and their combination is essential for remote sensing OVSS.

\subsection{Performance on Specialized Images}
We evaluate \samov on the GF-series datasets~\citep{chen2022introduction} to test its performance on Chinese  satellite images. As shown in Table~\ref{}, \samov performs well on target-specific datasets. On the GF-7 Building and Low-Grade Road datasets, it achieves 58.7\% and 60.2\% IoU, respectively, outperforming previous CLIP-based methods (\textit{e.g.}, ProxyCLIP at 34.9\% and 22.1\%).
However, on the GID dataset, the performance of \samov (42.2\% mIoU) is lower than that of SegEarth-OV (46.3\% mIoU). The GID dataset consists of GF-2 satellite images, which have a relatively lower spatial resolution and distinct imaging characteristics compared to UAV or sub-meter commercial satellite images. This resolution difference introduces a notable domain gap for SAM 3, as the model is primarily trained on natural images with sharp object boundaries. Consequently, SAM 3 may struggle with the blurred textures and continuous macroscopic land-cover categories (\textit{e.g.}, forest and farmland) in GF-2 images. In contrast, CLIP-based methods like SegEarth-OV benefit from extensive pre-training on diverse web data, providing stronger semantic generalization for such low-resolution or continuous scenes. Nevertheless, for precise object extraction in higher-resolution images, \samov remains a strong baseline.

\subsection{Performance on OVCD Task}

We evaluate the OVCD performance on the WHU-CD, LEVIR-CD, and S2Looking datasets. For the data processing, we crop the original test images of LEVIR-CD and WHU-CD into patches of $512 \times 512$ pixels. For the S2Looking dataset, we use the original $1024 \times 1024$ images directly for testing. Table~\ref{tab:ovcd} presents the quantitative comparison between our method and other change detection algorithms. Our method, which combines \samov with the joint instance- and pixel-level verification strategy, achieves the highest performance across all three benchmarks. For example, it reaches 88.0\% F1 and 78.6\% IoU on WHU-CD, and 85.3\% F1 and 74.3\% IoU on LEVIR-CD.
We observe two main findings from the results in Table~\ref{tab:ovcd}. First, existing change detection architectures show consistent performance improvements when using \samov as their segmentation backbone. For instance, the F1 score of DynamicEarth on WHU-CD increases from 75.9\% to 79.7\%, and on LEVIR-CD from 66.7\% to 72.0\%~\citep{su2026seg2change}. This demonstrates the high quality of the segmentation masks extracted by our \samov. Second, our proposed joint instance- and pixel-level verification strategy outperforms other matching algorithms (\textit{e.g.}, SFID and CACH) when using the same \samov foundation. This confirms that combining instance-level topology with the pixel-level comparison effectively reduces the impact of recognition errors and spatial misalignments. It enables robust zero-shot change detection without requiring any temporal training.

\subsection{Generalization to 3D Point Cloud Segmentation}

To extend \samov to 3D point cloud segmentation, we employ a multi-view projection strategy. Given a 3D point cloud and the corresponding camera intrinsic and extrinsic parameters, we project the 3D points onto multiple 2D image planes. We process these 2D images using \samov to obtain semantic predictions, which are then projected back to the 3D points. Because a single 3D point may be visible in multiple overlapping camera views, the 2D predictions can sometimes be inconsistent due to occlusions or varied perspectives. To aggregate these predictions and determine the final semantic label for each point, we apply a straightforward class-weighted voting mechanism.

We evaluate this approach on the STPLS3D-WMSC dataset. As shown in Table~\ref{tab:3d_evaluation}, our training-free method achieves a 61.26\% mIoU. It outperforms several fully supervised 3D models (\textit{e.g.}, PointTransformer, RandLA-Net, MinkowskiNet) trained on real and synthetic 3D data. This result demonstrates that the 2D visual representations from \samov are highly robust across different viewpoints. By combining these 2D predictions with a simple weighted voting strategy, the model identifies 3D structures effectively without relying on 3D point cloud annotations or complex geometric refinements.
In Figure~\ref{fig:3d}, we provide a qualitative visualization of a 3D UAV scene. The result shows that our method accurately categorizes distinct targets in a complex terrain. The predicted point cloud closely matches the ground truth, confirming that the multi-view projection strategy successfully translates 2D semantic knowledge into 3D spatial topologies.

\subsection{Performance on General Scene Datasets}

We benchmark \samov on three standard general scene datasets: Pascal VOC20, COCO Stuff, and Cityscapes (Table~\ref{tab:general}). For Pascal VOC20 and Cityscapes, we refine category names to better align with visual concepts; for instance, the ``terrain'' class in Cityscapes is expanded to ``grass, horizontal vegetation, soil, sand'' based on official definitions\footnote{https://www.cityscapes-dataset.com/dataset-overview}. \samov achieves dominance across all benchmarks. On Pascal VOC20, it achieves 96.8\% mIoU, surpassing both the best training-free method CorrCLIP (91.8\% mIoU) and the training-based CAT-Seg (94.6\% mIoU). On COCO Stuff, it reaches 42.8\% mIoU, significantly exceeding CorrCLIP by +8.8\%. The most significant improvement is observed on Cityscapes, where our method achieves 69.7\% mIoU, representing an increase of 18.6\% mIoU over the previous best result. These results further emphasize the powerful capabilities of SAM 3 and the effectiveness of \samov.

\section{Conclusion}


This paper explores the application of SAM 3 for open-vocabulary remote sensing interpretation. We introduce \samov, a training-free framework that adapts the prompt-based architecture of SAM 3 for complex geospatial scenarios. For semantic segmentation, we fuse the semantic and instance outputs and apply presence-guided filtering to address the scale variations and category sparsity inherent in Earth observation images. Building upon these fused predictions, we extend the framework to OVCD. By introducing a joint instance- and pixel-level verification strategy, the method effectively mitigates errors caused by spatial misalignments and recognition inconsistencies across bi-temporal images. Furthermore, we demonstrate the generalization capability of \samov by applying it to 3D point cloud segmentation via multi-view projection. Our comprehensive evaluation across 24 diverse remote sensing benchmarks shows that the proposed method establishes a strong training-free baseline for multi-dimensional remote sensing tasks. In certain scenarios, it even outperforms fully supervised models. These results confirm that, with appropriate adaptation mechanisms, general visual foundation models can serve as effective tools for open-vocabulary remote sensing analysis without requiring domain-specific training or complex architectures.


\footnotesize{\bibliography{library}}


\end{document}